\documentclass[letterpaper, 10 pt, conference]{ieeeconf}  

\IEEEoverridecommandlockouts                              

\usepackage{mathptmx} 
\usepackage{times} 
\usepackage{amsmath} 
\usepackage{amssymb}  

\usepackage{booktabs}
\usepackage{multirow}
\usepackage{graphicx}
\usepackage{url}
\usepackage{algorithm}
\usepackage{algpseudocode}

\title{\LARGE \bf
A Progress-Aware Leader-Follower Midair Docking System for Dual-Drone Aerial Manipulation
}
\author{Yifan Cai, Jan Ming Kevin Tan, Xiangqi Li, Chenzhe Jin, Narsimlu Kemsaram, and Valerio Modugno
\thanks{
Yifan Cai ({\tt\small y.cai.24@ucl.ac.uk}), Jan Ming Kevin Tan ({\tt\small jan.tan.24@ucl.ac.uk}), Xiangqi Li ({\tt\small xiangqi.li.24@ucl.ac.uk}), Chenzhe Jin ({\tt\small chenzhe.jin.23@ucl.ac.uk}), and Valerio Modugno ({\tt\small v.modugno@ucl.ac.uk}) are with the Department of Computer Science, University College London, London, United Kingdom.
}%
\thanks{
Narsimlu Kemsaram ({\tt\small narsimlu.kemsaram@um.edu.my}) is with the Department of Artificial Intelligence, University of Malaya, Kuala Lumpur, Malaysia.
}%
}

\begin{document}

\maketitle

\thispagestyle{empty}
\pagestyle{empty}


\begin{abstract}


Reliable midair docking between small unmanned aerial vehicles (UAVs) is essential for modular aerial cooperation and manipulation, but it requires precise relative-pose control and repeatable platform under tight thrust and payload constraints. We present a dual-drone docking platform where two quadrotors operate in a leader-follower formation and dock using a lightweight modular frame with passive magnetic latching. A progress-aware mission supervisor manages phase transitions: approach, alignment, capture, and settle. This platform integrates a complete hardware-software stack (ROS 2 with Crazyflie/PX4 interfaces) and synchronized logging for benchmark evaluation. We evaluate the platform in simulation and real-world experiments using quantitative metrics such as formation error, baseline and yaw consistency, docking success rate, time-to-dock, and failure-mode statistics. The platform enables statistically grounded comparison of docking supervision and synchronization strategies and provides a practical testbed for modular aerial cooperation and repeatable midair aerial manipulation.

\end{abstract}

\section{INTRODUCTION}


Reliable midair docking between small unmanned aerial vehicles (UAVs) is a foundational capability for modular aerial cooperation, in-flight reconfiguration, and cooperative manipulation. If two vehicles can repeatedly approach, align, and physically engage in flight, then a range of automation tasks becomes feasible, including self-assembly of larger structures, payload handover, and post-docking cooperative object transport. Prior work on modular aerial self-assembly (e.g., ModQuad) demonstrates the feasibility and importance of repeatable docking primitives for reconfigurable aerial systems \cite{saldana2018modquad}, \cite{li2019modquad}, \cite{gabrich2020modquad}. However, reliable docking remains difficult on micro-UAVs: small disturbances, sensing noise, and controller mismatch can quickly amplify into large relative errors near contact, causing failures such as bounce-off, misalignment, and timeout.


A key observation is that docking is not a single control problem but a multi-phase process. Successful engagement requires coordinated control of relative position, opposing yaw alignment, and closing speed during the final approach, followed by rapid damping of post-contact transients. In practice, robust docking therefore depends not only on low-level stabilization but also on mission-level supervision that (i) sequences the task into phases (approach, alignment, capture, and settle), (ii) enforces measurable tolerance gates before advancing phases, and (iii) maintains synchronized motion to suppress transient formation during coupled leader-follower maneuvers. These considerations align with broader findings in cooperative aerial manipulation, where systems-level coordination and constraints play a critical role in reliability \cite{michael2011cooperative}, \cite{mellinger2013cooperative}, \cite{yang2022collaborative}.

\begin{figure}[t]
    \centering
    \includegraphics[width=0.45\textwidth]{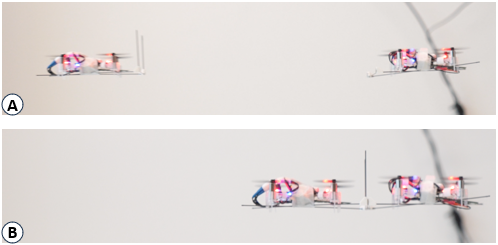}
    \caption{
    Proposed progress-aware leader-follower midair docking system for dual-drone aerial manipulation with two Crazyflie~2.1 drones: (A) approach phase where the follower converges to the leader under progress-aware supervision, and (B) latched docking configuration with stable alignment.
    }
    \label{fig:crazyflie_docking_ps}
\end{figure}

This paper presents a Crazyflie-based dual-drone midair docking platform for repeatable experimentation and benchmark-style evaluation. Two Crazyflie 2.1 micro-quadrotors carry a gram-scale modular magnetic frame that enables passive capture once alignment conditions are met. 
Figure \ref{fig:crazyflie_docking_ps} shows the Crazyflie~2.1 leader-follower approach and the achieved midair magnetic latch in real flights.
A motion-capture-driven state interface provides consistent position and yaw estimates for both vehicles, and the software stack is organized so the estimator-to-setpoint interfaces are transferable to PX4-class systems. Low-level tracking uses Proportional-Integral-Derivative (PID) outer loops with gains tuned offline via Bayesian Optimization (BO) to reduce overshoot and improve settling under limited thrust margins \cite{shahriari2015taking}, \cite{bulka2022experiments}, \cite{gedefaw2025review}. For cooperation, a progress-aware leader-follower supervisor issues synchronized setpoints, regulates a fixed baseline with opposing yaw, and advances mission phases only when both vehicles satisfy distance, yaw, and velocity tolerances, reducing premature contact and accordion transients during coupled motion.
Experiments in an indoor motion-capture arena and supporting simulation show a repeatable approach and docking behavior. Tracking remains stable, and capture transients are damped quickly.

The main contributions of this work are:
i) A progress-aware mission supervisor that formalizes phase gating for leader-follower docking using measurable guards,
ii) A lightweight modular docking platform compatible with micro-UAV payload constraints, enabling repeatable midair engagement,
iii) An integrated software pipeline (ROS 2 with Crazyflie/PX4 interfaces) with a synchronized measurement pipeline for dual-drone experiments, and
iv) An experimental evaluation in simulation and real-world environments to quantify docking repeatability.

This paper is organized as follows: 
Section II reviews modular aerial docking, leader-follower formation, stabilization, and synchronization.
Section III presents the concept, design, modeling, and methods: Crazyflie dual-drone platform, leader-follower formation, and docking mechanism. 
Section IV reports experiments, results, 
discussions, limitations, and future research directions. Finally, Section V concludes the paper.

\section{RELATED WORK}

Midair docking of small UAVs enables modular self-assembly, cooperative aerial manipulation, and in-flight reconfiguration. Reliable docking in the real world depends not only on low-level control but also on repeatable engagement geometry and capture mechanics, relative formation control tailored to the \textit{approach-align-capture} sequence, robust stabilization and data-driven tuning under tight thrust margins, and time synchronization for dual-drone coordination.



\subsection{Modular Aerial Docking and Self-Assembly}

Modular aerial robots seek to assemble structures in flight and therefore rely on docking interfaces with passive alignment, capture tolerance, and sufficient stiffness for post-contact stability. The ModQuad series demonstrated midair self-assembly using lightweight structural frames and docking interfaces designed for repeatable engagement \cite{saldana2018modquad}. Related efforts in aerial self-assembly and docking similarly emphasize mechanical design choices (lead-in geometry, capture, alignment, and latching) as key drivers of docking reliability, often as much as the controller \cite{li2019modquad}. Beyond self-assembly, cooperative aerial manipulation has been widely studied for contact-based transport and grasping \cite{michael2011cooperative}, \cite{mellinger2013cooperative}, emphasizing coordination, safety constraints, and disturbance rejection \cite{gandhi2020self}, \cite{gabrich2021flying}. In contrast, our work focuses on the pre-assembly stage. We formalize a repeatable docking workflow with clear success criteria, failure modes, and benchmarking metrics.

\subsection{Leader-Follower Formation for Docking}

Leader-follower control is widely used for multi-UAV coordination, where a leader tracks global waypoints and followers maintain formation \cite{liu2024leader}, \cite{wang2025distributed}. In docking tasks, however, leader-follower waypoint tracking alone is often inadequate \cite{chen2018improved}, \cite{li2025resilient}. The system must handle phase transitions, including approach, alignment, capture, and settle, while avoiding transient amplification and enforcing safety when vehicles are mechanically coupled. Previous cooperative manipulation systems indicate that stability and safety rely more on robust supervisory logic and constraints than on new low-level controllers \cite{incremona2013supervisory}, \cite{wu2021distributed}. Our platform uses a leader-follower structure with an added progress-aware supervisor. This supervisor incorporates a fixed baseline with opposing yaw, synchronized setpoints, and progress gating. It advances mission phases only when both drones meet specific distance, yaw, and velocity tolerances. This method improves docking repeatability and enables controlled ablation studies, such as supervisor ON/OFF, compared to a waypoint-only leader-follower baseline.


\subsection{Stabilization and Data-Driven Gain Tuning} 

Micro-UAVs operate with limited thrust margins and are sensitive to model mismatch, sensor noise, and aerodynamic disturbances. As a result, classical PID or outer-loop tracking remains competitive when gains are carefully tuned \cite{giernacki2017crazyflie}, \cite{preiss2017crazyswarm}, \cite{pichierri2023crazychoir}. Data-driven tuning methods, including Bayesian optimization, have been widely used to tune controller parameters as a black-box optimization problem with few trials, balancing exploration and exploitation while respecting safety constraints \cite{paulson2020data}, \cite{sorourifar2021data}, \cite{khosravi2021performance}. For docking, improved damping, reduced overshoot, and lower velocity at contact directly increase capture probability and reduce bounce-off and misalignment failures. In our work, Bayesian optimization-tuned gains are an enabling component for repeatable experimentation and fair benchmarking: we report quantitative improvements (settling time/overshoot, RMS errors) and include fixed-gain ablations.



\subsection{Time Synchronization in Multi-UAV Systems} 

Time synchronization among vehicles and sensing infrastructure is critical for consistent state estimation, coordinated phase transitions, and the generation of comparable experimental logs \cite{khan2023ieee}. 
Most practical multi-UAV systems achieve millisecond to sub-millisecond alignment over onboard networks, which is sufficient for formation waypoints and data fusion. However, this level of synchronization is inadequate for payloads that require phase coherence at high frequencies. For timing-critical manipulation payloads, microsecond-level skew is essential. For instance, at 40 kHz, one acoustic period is approximately 25 microseconds. Thus, multi-microsecond skew alters the relative phase and can degrade cooperative interference. Standards-based synchronization, such as IEEE 1588 Precision Time Protocol (PTP), can achieve microsecond-level alignment, particularly over wired connections \cite{khan2023ieee}. 
Ultra-wideband (UWB) systems can also support precise timing in GPS-denied environments when combined with calibration and drift compensation \cite{kannisto2005software}, \cite{cano2019kalman}, \cite{adriaens2025high}. In laboratory settings, motion-capture systems offer a shared clock and high-accuracy pose streams, facilitating reproducible benchmarking, albeit with reliance on specific infrastructure.

This work presents our contribution as a reproducible docking-and-synchronization platform suitable for integrating timing-sensitive aerial manipulation payloads, such as contactless cooperative acoustic aerial manipulation tasks.

\section{CONCEPT, DESIGN, MODELING, and METHODS}

This section presents the proposed dual-drone platform for repeatable leader-follower docking under a motion-capture (MoCap) system. The system is designed as a lightweight experimental testbed that enables controlled evaluation of relative-pose regulation, synchronized formation motion, and midair engagement within the limited payload and thrust margins of micro-UAVs. The overall method combines five tightly coupled components: (i) a Crazyflie-based dual-drone platform, (ii) a layered software-hardware architecture for sensing, state estimation, and control, (iii) a leader-follower formation model that generates synchronized docking setpoints, (iv) a progress-aware mission supervisor that governs phase transitions, and (v) a lightweight modular docking mechanism that enables repeatable capture.

\subsection{Crazyflie-Based Dual-Drone System}

\begin{figure*}[!htbp]
    \centering
    \includegraphics[width=0.60\textwidth]{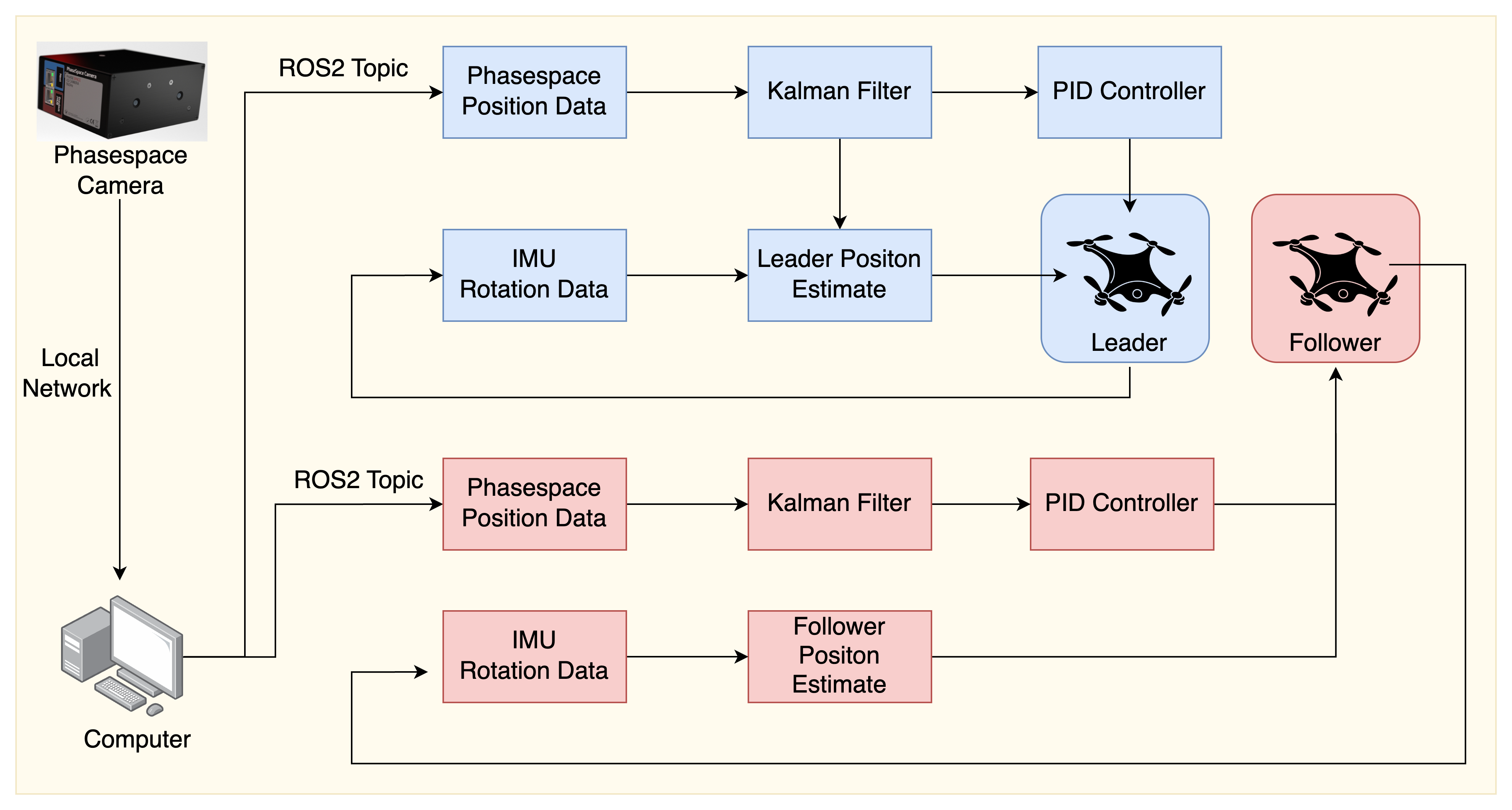}
    \caption{
    Crazyflie-based dual-drone docking system under PhaseSpace motion capture: MoCap/IMU fusion with per-vehicle EKF, progress-aware leader-follower coordination, and lightweight magnetic frame enabling repeatable engagement.
    }
    \label{fig:SystemOverview_Crazyflie}
\end{figure*}

The proposed platform uses two Crazyflie 2.1 micro-quadrotors in an indoor PhaseSpace motion-capture arena. The Crazyflie platform is chosen for rapid iteration, low-risk experimentation, and repeatable testing under tight payload constraints \cite{giernacki2017crazyflie}, \cite{preiss2017crazyswarm}. PhaseSpace provides high-rate pose measurements for both vehicles in a shared global frame \cite{phasespace2020}. Each Crazyflie also streams onboard IMU measurements (e.g., angular rates) to the ground computer through the CRTP radio link. 
Figure \ref{fig:SystemOverview_Crazyflie} summarizes the end-to-end docking pipeline, including MoCap/IMU fusion, state estimation, and progress-aware coordination.

For each drone, a Kalman filter (implemented as an EKF in our stack) fuses PhaseSpace measurements with IMU-derived motion information to estimate the reduced state:
\begin{equation}
\mathbf{x}_i=[x_i,\;y_i,\;z_i,\;\dot{x}_i,\;\dot{y}_i,\;\dot{z}_i,\;\psi_i]^\top
\label{eq:state}
\end{equation}
where, \(i\in\{L,F\}\) denotes the leader and follower, respectively. 
The leader state is broadcast to the follower so that relative control can be computed consistently from estimated states (rather than raw pose streams) \cite{kalman1960new}, \cite{maybeck1979new}.
These estimated states are used for all subsequent control and docking decisions.

\subsection{Dual-Drone System Architecture}

\begin{figure}[!htbp]
    \centering
    \includegraphics[width=0.25\textwidth]{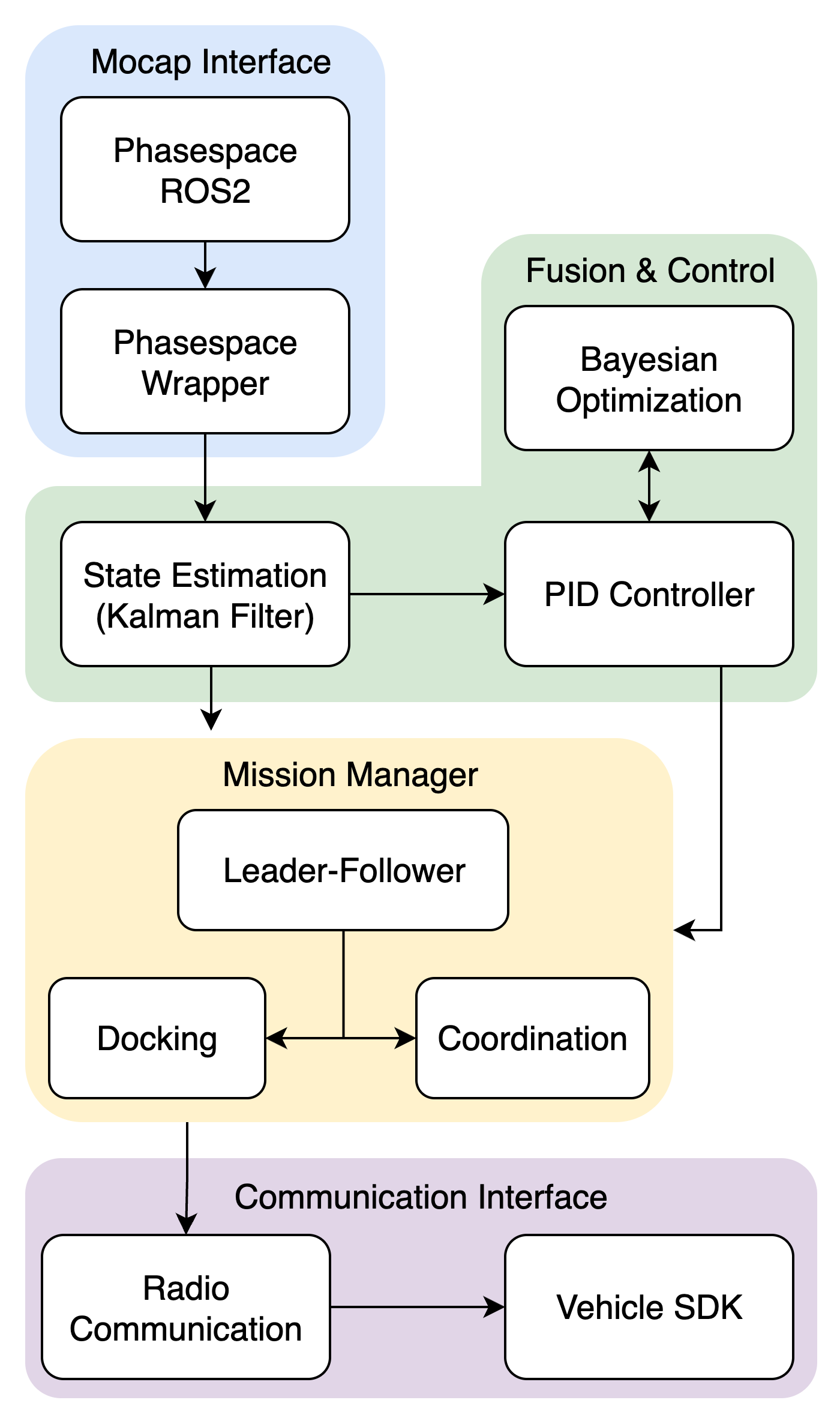}
    \caption{
    Software architecture for dual-Crazyflie docking: MoCap/IMU sensing, per-vehicle EKF state estimation, BO-tuned PID outer-loop tracking, and a progress-aware mission supervisor for phase gating and safety.
    }
    \label{fig:SoftwareArchitecture}
\end{figure}

The software architecture is organized as a layered pipeline consisting of sensing, state estimation, control, mission supervision, and communication. 
The layered software components and information flow used in our experiments are depicted in Figure~\ref{fig:SoftwareArchitecture}.

\subsubsection{Sensing layer}

The PhaseSpace system publishes drone poses as ROS 2 topics, while onboard IMU measurements are streamed through the Crazyflie radio link \cite{pichierri2023crazychoir}, \cite{phasespace2020}. All streams are stamped with a common host clock, which serves two purposes: first, it guarantees consistent replay and post-hoc analysis of docking trajectories, and second, it provides the synchronization reference required by the progress-aware supervisor.

\subsubsection{State estimation layer}


An EKF runs per drone to estimate \(\{\mathbf{p},\mathbf{v},\psi\}\) using MoCap and IMU inputs \cite{kalman1960new}, \cite{maybeck1979new}. 

\subsubsection{Control layer}

PID outer-loop controllers generate setpoints for each drone. We tune gains using Bayesian optimization offline to reduce overshoot and settling time under tight thrust margins \cite{gedefaw2025review}, \cite{shahriari2015taking}.

\subsubsection{Mission supervision}




A progress-aware mission supervisor manages phase transitions (\textit{Approach \(\rightarrow\) Align \(\rightarrow\) Capture \(\rightarrow\) Settle}) and enforces tolerance gates before advancing phases. 

\subsubsection{Communication and safety}


Commands and telemetry are exchanged via the Crazyflie radio link (CRTP). Safety includes watchdog timeouts, hover fallback, and emergency-stop logic.

\subsection{Leader-Follower Formation Model}

Let the leader and follower positions be,
\begin{equation}
\label{eq:LFP}
\mathbf{p}_L=[x_L,y_L,z_L]^\top,\qquad
\mathbf{p}_F=[x_F,y_F,z_F]^\top
\end{equation}
and let the operator specify the desired formation center,
\[
\mathbf{g}=[x^\ast,y^\ast,z^\ast]^\top
\]
The drones are assigned a fixed docking separation \(d_{\mathrm{dock}}\) along the real-world \(x\)-axis, with baseline vector,
\[
\mathbf{b}=[d_{\mathrm{dock}},0,0]^\top
\]
The synchronized formation targets are then,
\begin{equation}
\mathbf{p}_L^\star=\mathbf{g}-\frac{1}{2}\mathbf{b}, \qquad
\mathbf{p}_F^\star=\mathbf{g}+\frac{1}{2}\mathbf{b}
\label{eq:lf_targets_case_short}
\end{equation}
with fixed yaw references,
\[
\psi_L^\star=0,\qquad \psi_F^\star=\pi
\]
This opposing-yaw convention enforces a face-to-face docking geometry during approach.

To reduce accordion-type effects, the synchronized cruise speed is chosen conservatively from the two vehicle capabilities,
\begin{equation}
v_{\mathrm{sync}}=0.8\min\!\left(v_{\mathrm{form}}^{(L)},v_{\mathrm{form}}^{(F)}\right)
\label{eq:vsync_case_short}
\end{equation}
and the timeout is derived from the commanded formation-center displacement,
\begin{equation}
\hat{t}=\frac{\|\mathbf{g}-\mathbf{c}\|}{v_{\mathrm{sync}}}, \qquad
t_{\max}=\mathrm{clip}(2\hat{t},30~\mathrm{s},t_{\mathrm{usr}})
\label{eq:timeout_case_short}
\end{equation}
where, \(\mathbf{c}=(\mathbf{p}_L + \mathbf{p}_F)/2\) is the current formation center.

\subsection{Progress-Aware Docking Method}
A direct leader-follower controller is often insufficient for repeatable docking because small velocity mismatches or yaw deviations can cause bounce-off or incomplete engagement at contact. To address this, the proposed method augments leader--follower tracking with explicit docking phases and gating conditions (see Algorithm \ref{alg:progress_aware_docking}).

\begin{algorithm}[!htbp]
\caption{Progress-aware leader-follower docking}
\label{alg:progress_aware_docking}
\begin{algorithmic}[1]
\State \textbf{Input:} target center \(\mathbf{g}\), timeout \(t_{\mathrm{usr}}\)
\State Compute \(\mathbf{p}_L^\star,\mathbf{p}_F^\star\) from \eqref{eq:lf_targets_case_short}
\State Compute \(v_{\mathrm{sync}}\) and \(t_{\max}\) from \eqref{eq:vsync_case_short}--\eqref{eq:timeout_case_short}
\State Command both drones with synchronized position targets and fixed yaw
\State Enter \textsc{Approach}
\While{mission active}
    \State Estimate baseline, yaw errors and relative speed from EKF states, \(e_b,e_\psi,v_{\mathrm{rel}}\)
    \If{\textsc{Approach} and baseline error within coarse tolerance (\(|e_b|<\epsilon_b^{(1)}\))} enter \textsc{Align} \EndIf
    \If{\textsc{Align} and \(|e_b|<\epsilon_b^{(2)}\), \(|e_\psi|<\epsilon_\psi\), \(v_{\mathrm{rel}}<\epsilon_v\)} enter \textsc{Capture} \EndIf
    \If{\textsc{Capture} and latch detected} enter \textsc{Settle} \EndIf
    \If{\textsc{Settle} and hold time \(>T_{\mathrm{hold}}\)} declare success \EndIf
    \If{timeout or safety fault} abort to hover or land \EndIf
\EndWhile
\end{algorithmic}
\end{algorithm}

During \textit{Approach}, the follower is guided toward a coarse docking corridor centered on the desired baseline. During \textit{Align}, tighter tolerances are imposed on relative yaw and closing speed. During \textit{Capture}, contact is allowed only if the follower approaches within the alignment window at low relative velocity. Finally, during \textit{Settle}, the docked pair must remain attached for a minimum hold time while oscillations decay below a fixed threshold.

These gating conditions are evaluated from filtered relative states,
\begin{equation}
e_b = \left\|\mathbf{p}_F-\mathbf{p}_L\right\|-d_{\mathrm{dock}}, \qquad
e_\psi = (\psi_F-\psi_L)-\pi
\end{equation}
together with the relative speed \(v_{\mathrm{rel}}\). A phase transition is permitted only when,
\begin{equation}
|e_b|<\epsilon_b,\qquad |e_\psi|<\epsilon_\psi,\qquad v_{\mathrm{rel}}<\epsilon_v
\end{equation}
where, \(\epsilon_b\), \(\epsilon_\psi\), and \(\epsilon_v\) are design tolerances. This simple gating mechanism converts docking into a measurable, benchmarkable process with clearly defined success conditions.

\subsection{Mechanical Docking Design (Modular Frame)}

\begin{figure}[!htbp]
    \centering
    \includegraphics[width=0.40\textwidth]
    {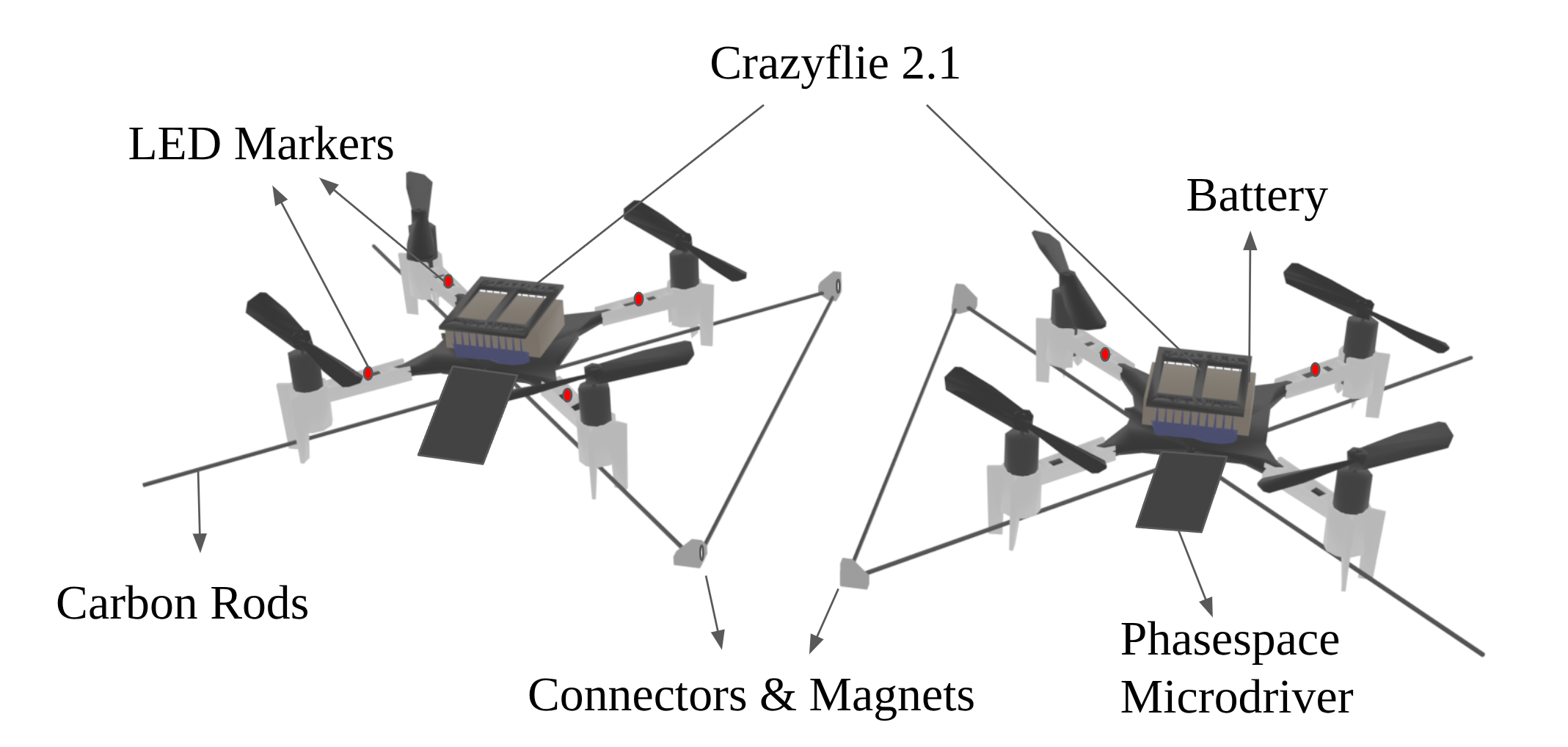}
    \caption{Crazyflie~2.1 platform with modular docking frame: LED markers and PhaseSpace microDriver for tracking, carbon-fiber rods for rigidity, 3D-printed connectors with embedded magnets for docking, and onboard battery.}
  \label{fig:crazyflie_structure}
\end{figure}

\begin{figure}[!htbp]
    \centering
    \includegraphics[width=0.40\textwidth]
    {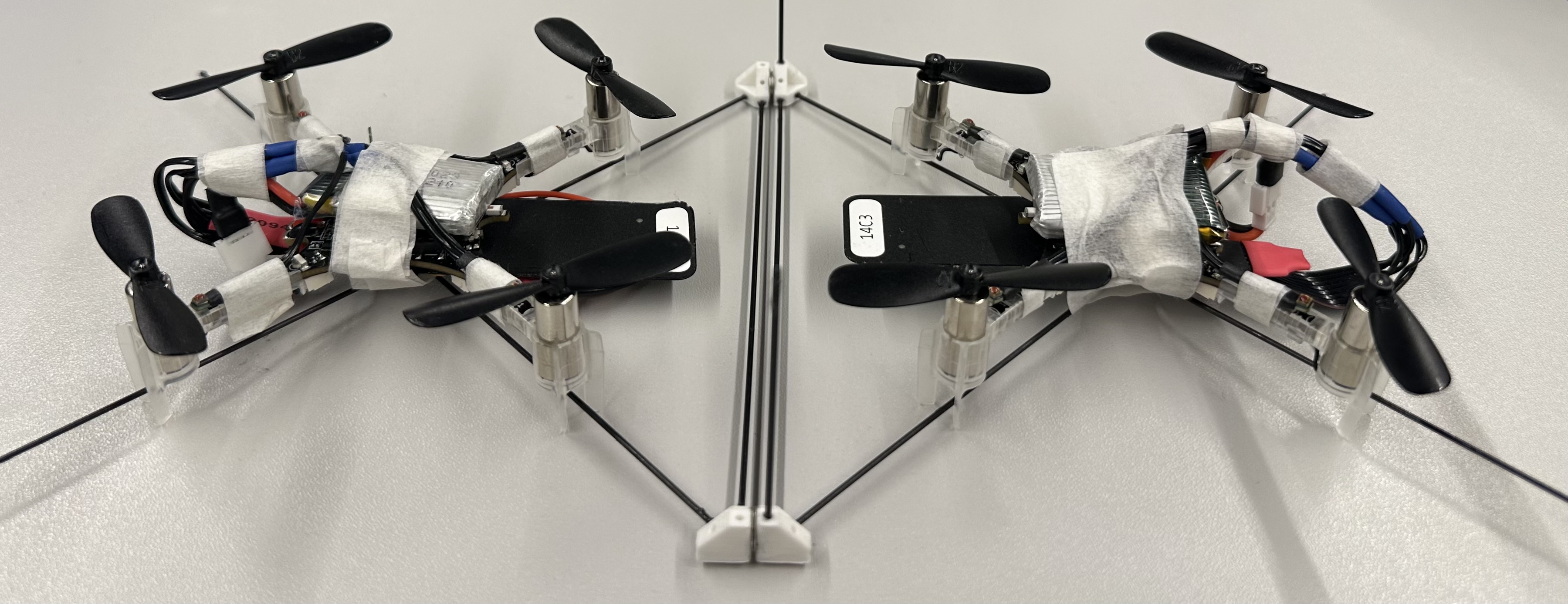}
    \caption{Experimental hardware for dual-drone docking: two Crazyflie~2.1 micro-quadrotors with carbon-fiber rod frames and 3D-printed connectors providing the modular docking interface used in flight trials.}
  \label{fig:crazyflie_docking}
\end{figure}

To improve repeatability at contact while respecting payload limits, each Crazyflie is enclosed in a lightweight modular frame constructed from carbon-fiber rods and 3D-printed connectors. 
The modular docking frame and its key hardware elements are illustrated in Figure \ref{fig:crazyflie_structure}, while Figure \ref{fig:crazyflie_docking} shows the assembled hardware used in experiments.

Carbon-fiber rods define the outer structure, while 3D-printed connectors join the rods and house embedded magnets. The geometry of the connector provides passive alignment during contact, distributing the docking load into the frame rather than into the flight electronics. Because the Crazyflie payload margin is approximately $15g$, all structural elements are designed to minimize mass while maintaining sufficient stiffness for repeatable engagement.
The magnetic connectors are used here as passive actuators for capture. Their role is to increase docking repeatability once the relative pose and speed conditions are met. This allows the experimental platform to isolate the upstream problems of formation control, synchronization, and low-speed engagement from the downstream problem of rigid post-contact attachment.

\section{EVALUATION and RESULTS}\label{sec:EvaluationResults}

This section evaluates the proposed dual-drone docking platform using controlled motion-capture experiments with two Crazyflie 2.1 micro-UAVs. The goal is to quantify (i) synchronized leader-follower motion quality, (ii) repeatability of the progress-aware docking procedure, and (iii) effect of supervisory gating and controller tuning on docking outcomes. We report metrics including docking success rate, time-to-dock, baseline/yaw consistency, and failure modes.

\subsection{Experimental Setup}

\begin{figure}[!htbp]
    \centering
    \includegraphics[width=0.40\textwidth]
    {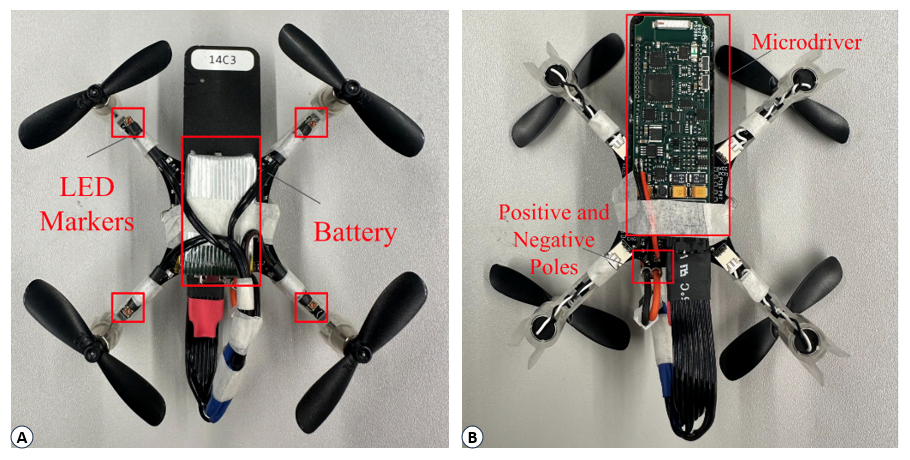}
    \caption{Crazyflie hardware for docking validation: modular magnetic frame and tracking instrumentation. (A) front view and (B) back view showing microDriver, battery, and magnet layout.}
  \label{fig:crazyflie_hardware}
\end{figure}

The Crazyflie 2.1 pair is used as a lightweight platform for cooperative docking. Each drone carries its stock flight board, battery, LED markers, and PhaseSpace microDriver for motion capture. A modular magnetic frame made of carbon-fiber rods with 3D-printed connectors and embedded NdFeB magnets enables repeatable mid-air engagement within the $\sim$15\,g payload envelope. A 2.4 GHz radio dongle provides the CRTP link for commands and telemetry (see Figure \ref{fig:crazyflie_hardware}). 

\subsubsection{Arena and sensing} 

Real-world experiments are conducted in an indoor \(6 \times 6\)~m PhaseSpace motion-capture arena. Each Crazyflie carries LED markers and microDrivers for tracking. The MoCap system tracks each Crazyflie and publishes their poses to a ground computer as ROS 2 topics. Each Crazyflie also streams onboard IMU data via the CRTP radio link. ROS~2 nodes on the ground computer fuse MoCap and IMU in a per-vehicle Kalman filter to estimate position, velocity, and yaw for both leader and follower. All messages are timestamped with a host-synchronized time base to ensure consistent timing throughout the control loop and post-processing.


\subsubsection{Vehicles and configuration} 

The leader and follower fly at a nominal formation height \(z^\ast\) and maintain a fixed docking separation \(d^\star\) with opposing yaw references \((\psi_L^\star=0,\ \psi_F^\star=\pi)\). In our work, the docking distance is fixed at \(d^\star=0.46\)~m (matching the simulation baseline) to preserve rotor-disk clearance and provide a consistent engagement geometry across trials.


\subsubsection{Trial protocol} 

Each trial follows the sequence: (i) takeoff and stabilize at hover, (ii) enter leader-follower formation, (iii) execute the progress-aware docking routine (\textit{Approach \(\rightarrow\) Align \(\rightarrow\) Capture \(\rightarrow\) Settle}), (iv) hold the docked configuration for a fixed duration, (v) land, and (vi) reset.

\subsubsection{Success criteria and timing}

A trial is counted as \emph{successful} if the system reaches \textit{Settle} (latch confirmed) and remains attached for at least \(T_{\text{hold}}\) without triggering safety abort. \emph{Time-to-dock} is measured from the start of \textit{Approach} to the first entry into \textit{Settle}. Failure modes are categorized as timeout (no capture before \(T_{\text{dock}}\)), misalignment (yaw/baseline never meets gate), bounce-off (contact without sustained latch), or safety abort.

\subsection{Dual-Drone Evaluation in Simulation}

\begin{figure*}[!htbp]
  \centering
  \begin{minipage}[b]{0.40\textwidth}
    \includegraphics[width=\textwidth]{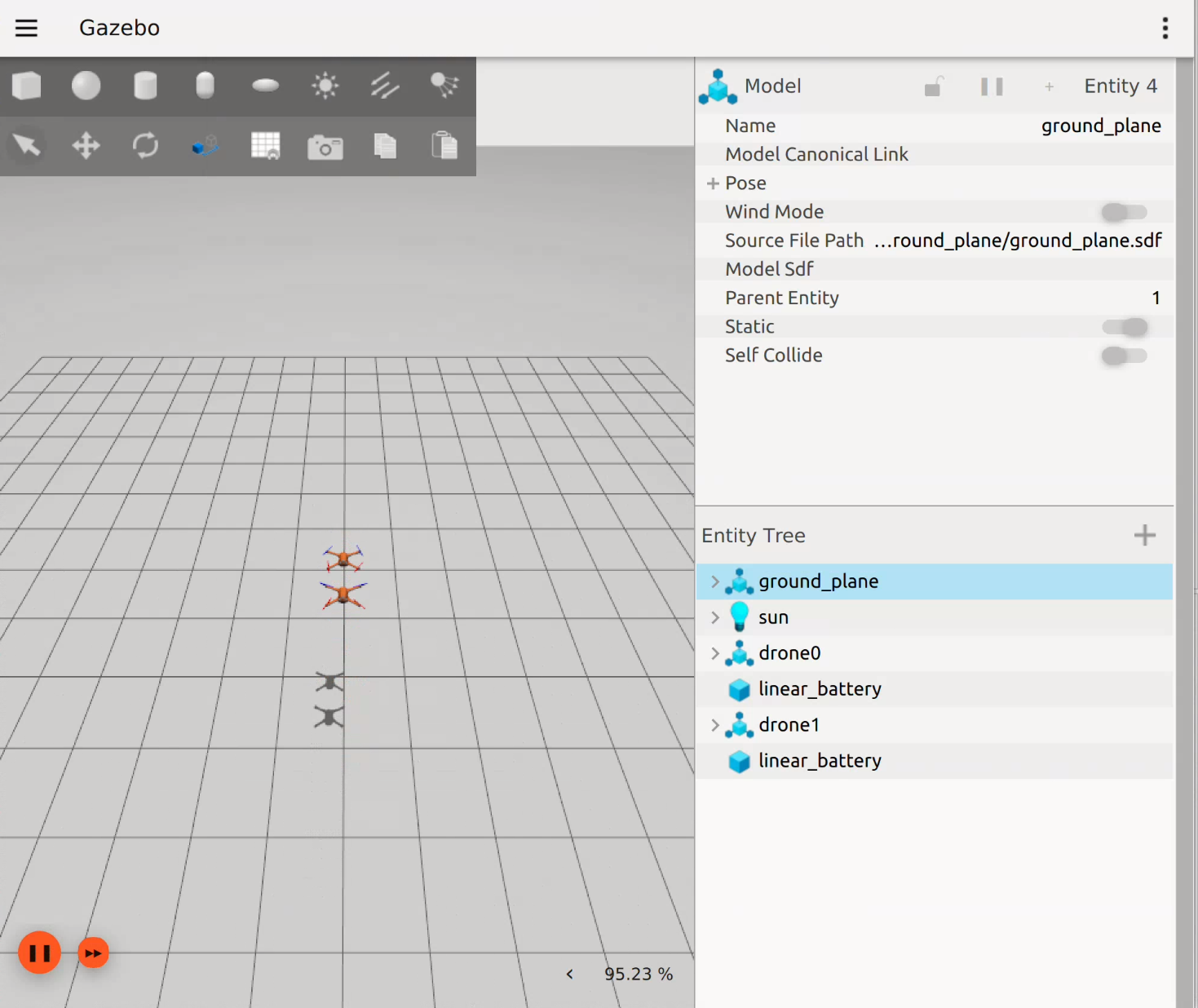}
  \end{minipage}
  \hfill
  \begin{minipage}[b]{0.40\textwidth}
    \includegraphics[width=\textwidth]{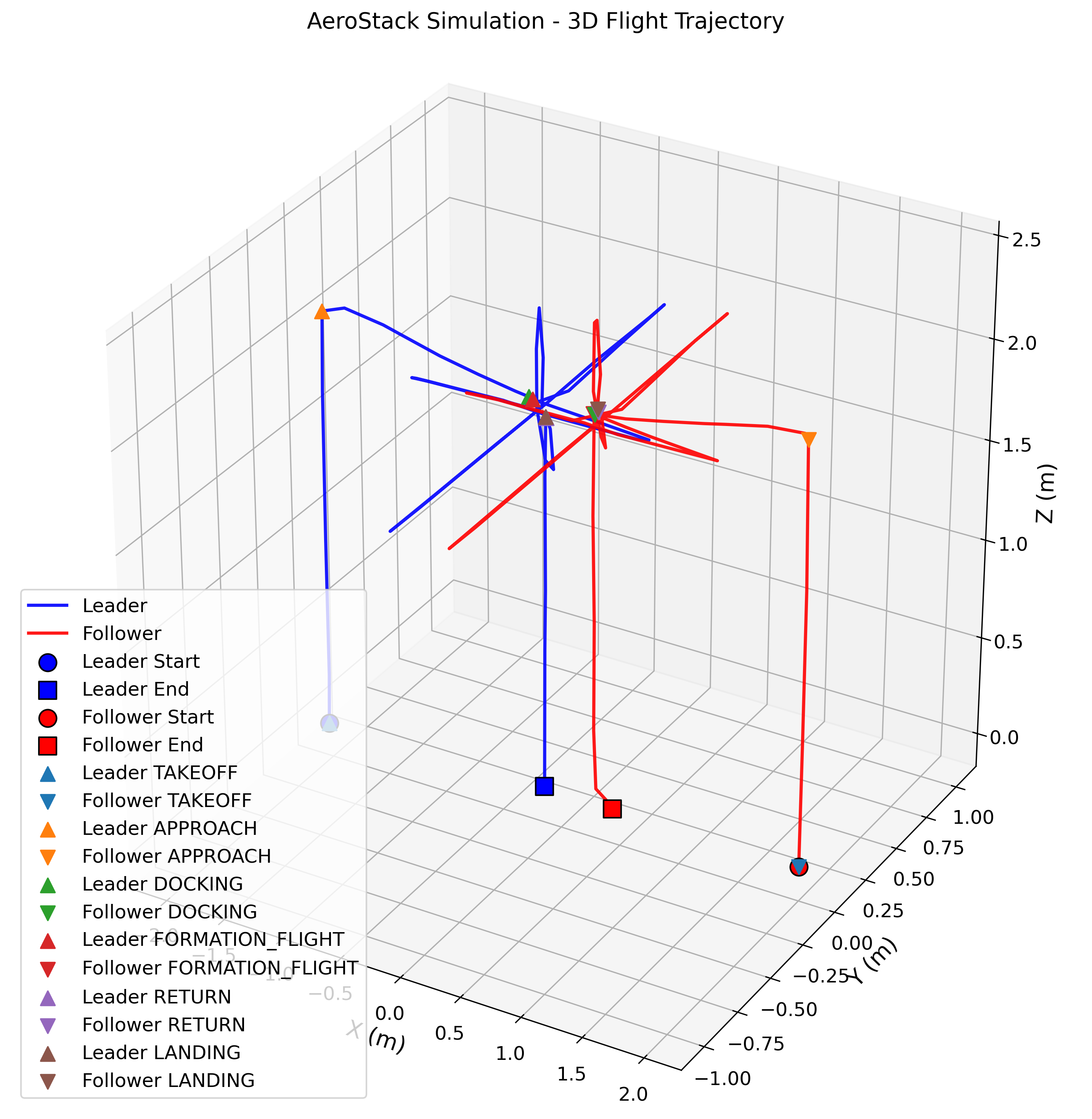}
  \end{minipage}
  \caption{Simulation in Gazebo/Aerostack2: Leader-follower trajectories showing convergence to a stable docking configuration and constant-baseline formation after docking.}
  \label{fig:simulation_results}
\end{figure*}

We first validate the mission supervisor and leader-follower policy in Aerostack2 + Gazebo simulation using a six-phase mission: \textit{takeoff \(\rightarrow\) approach \(\rightarrow\) dock \(\rightarrow\) formation \(\rightarrow\) return \(\rightarrow\) land}. 
Simulation results in Figure \ref{fig:simulation_results} show convergence to the docked configuration and constant-baseline formation. 
Both agents complete all phases without aborts. The 3D trajectories converge to a stable docking geometry and constant-baseline formation maneuvers. The measured baseline is \(0.46\)~m \(\pm 0.005\)~m with yaw alignment within \(5^\circ\). The docking distance is fixed at \(d^\star=0.46\)~m to maintain approximately \(0.10\)~m rotor-disk clearance and to enforce a consistent engagement geometry.
In simulation, altitude stabilizes around \(2.0\)~m \(\pm 0.05\)~m (note that this simulation altitude setpoint differs from the lower flight altitude used in real experiments for safety and tracking reliability).

\subsection{Dual-Drone Evaluation in Real-World}



Two Crazyflies execute \textit{takeoff \(\rightarrow\) approach \(\rightarrow\) dock \(\rightarrow\) formation \(\rightarrow\) return \(\rightarrow\) land} using the progress-aware leader-follower policy. The follower maintains a fixed baseline with opposing yaw during the docking window, and altitude holds near \(\sim 0.5\)~m with small oscillations that are rapidly damped. Short capture transients (e.g., follower \(v_z\) peaking near \(\sim 2\)~m/s) decay within a few seconds, yielding bounded velocity profiles and stable engagement.

\begin{figure}[!htbp]
    \centering
    \includegraphics[width=0.40\textwidth]{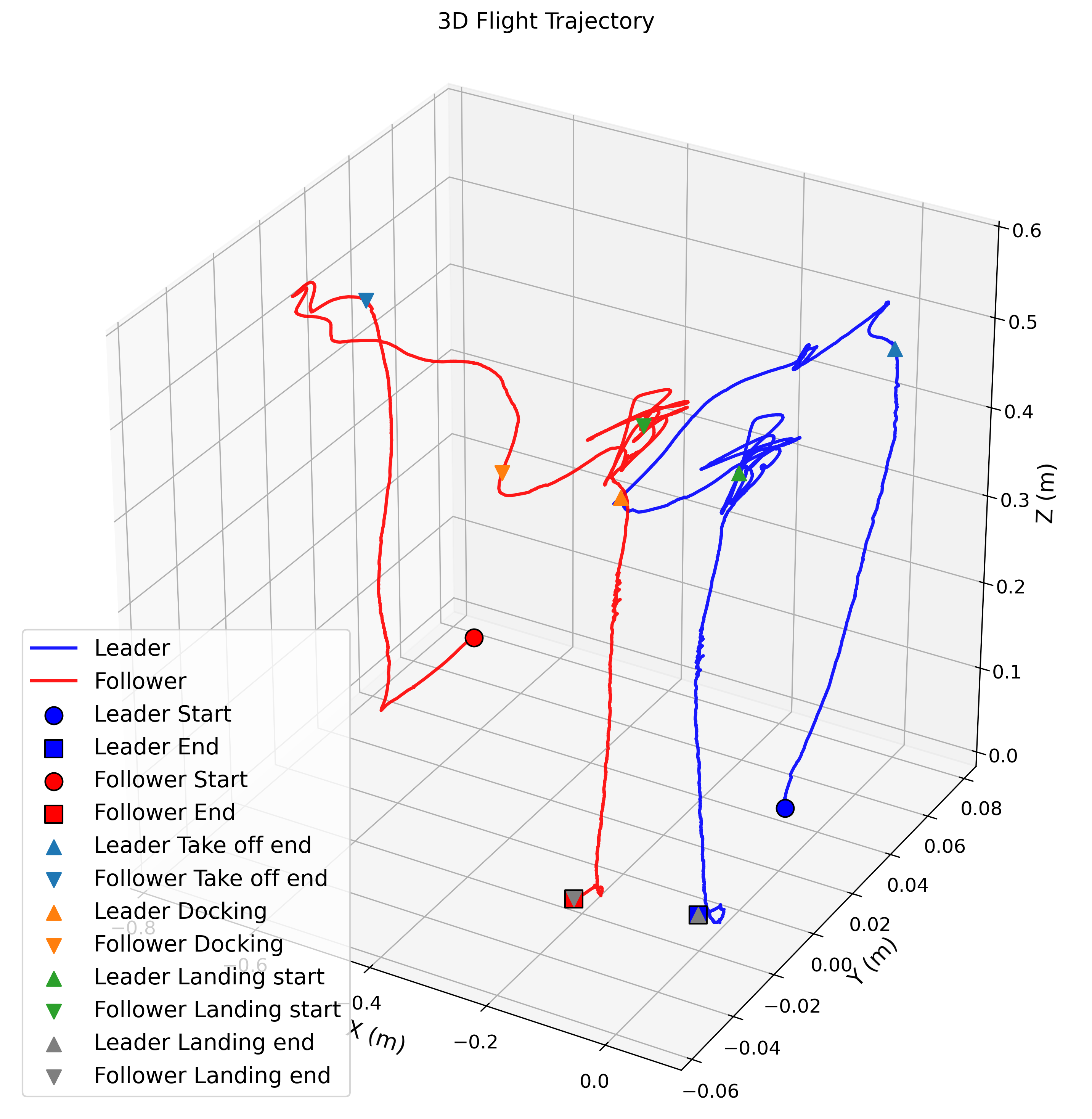}
    \caption{Real-world leader (blue) and follower (red) 3D trajectories from takeoff to docking and landing. The pair maintains a consistent baseline during the docking window.}
    \label{fig:crazyflie_trajectory}
\end{figure}
\begin{figure}[!htbp]
    \centering
    \includegraphics[width=0.40\textwidth]{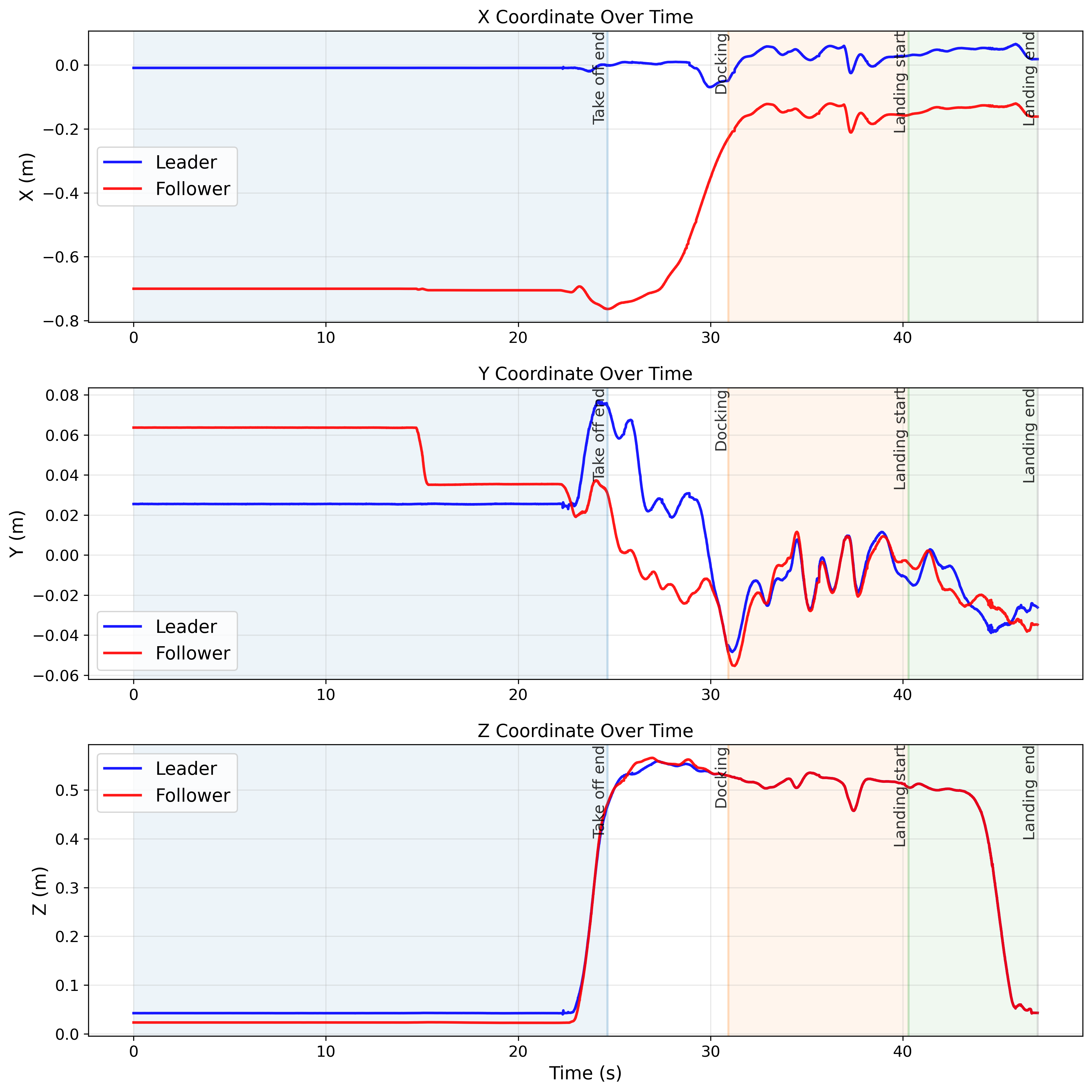}
    \caption{Dual-Crazyflie Cartesian position time histories \((x,y,z)\). Both vehicles converge to the flight altitude and maintain docking alignment, and capture-induced oscillations are rapidly suppressed.}
    \label{fig:crazyflie_xyz_icra}
\end{figure}
\begin{figure*}[!htbp]
    \centering
    \includegraphics[width=0.60\textwidth]{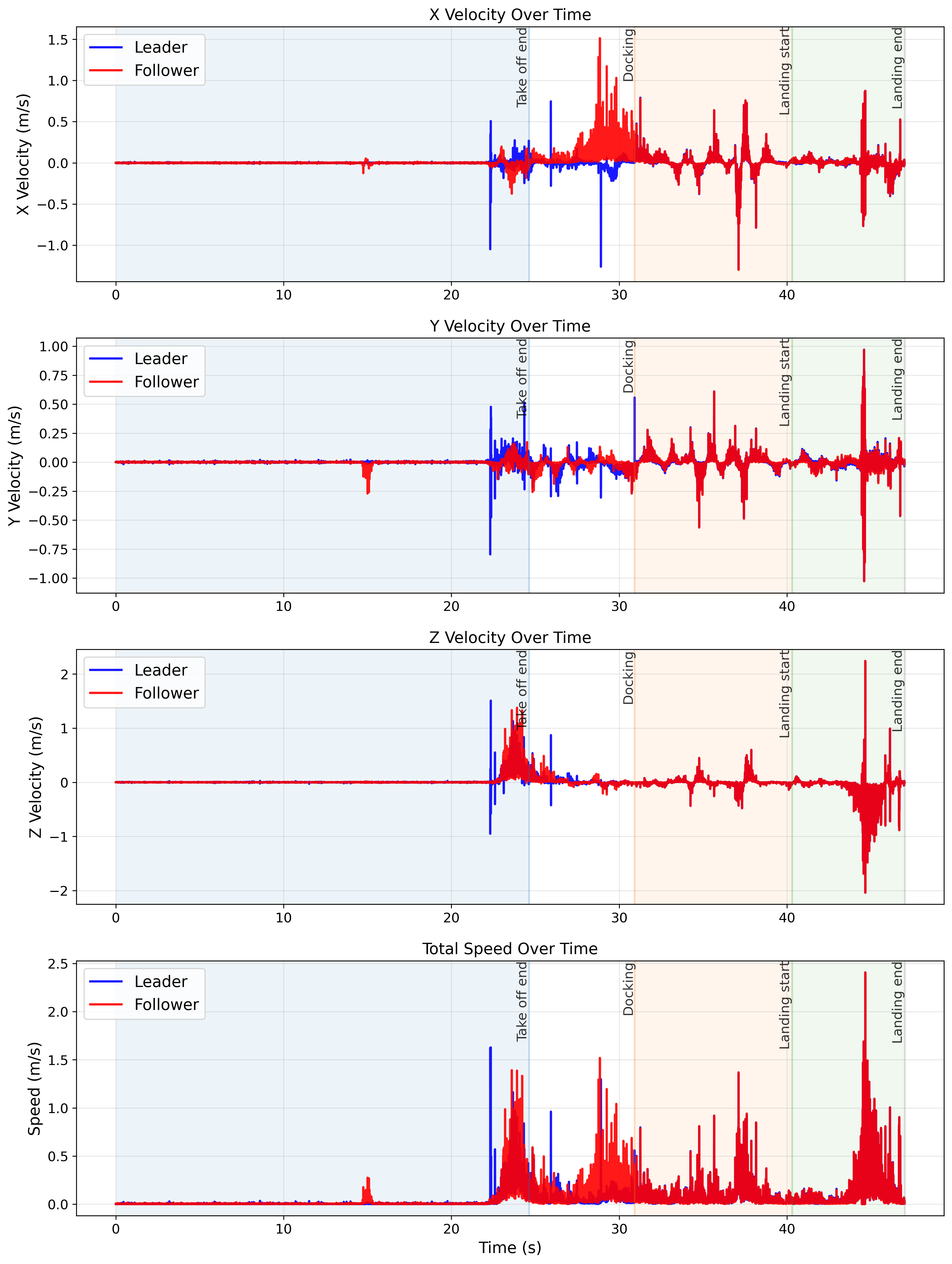}
    \caption{Dual-Crazyflie velocity profiles \((v_x,v_y,v_z)\) and total speed. Docking transients are rapidly damped, preserving synchronized motion across mission phases.}
    \label{fig:crazyflie_vel_icra}
\end{figure*}

Real-world 3D trajectories for the leader (blue) and follower (red) are shown in Figure \ref{fig:crazyflie_trajectory}, demonstrating consistent baseline regulation during docking.
Figure \ref{fig:crazyflie_xyz_icra} reports Cartesian position time histories, showing convergence to the flight altitude and bounded oscillations during capture.
After takeoff, both drones settle to \(\approx 0.5\)~m altitude and remain within the docking window in \(x\) and \(y\). The follower tightly tracks the leader with small capture oscillations that are quickly damped, supporting repeatable engagement.
Velocity time histories in Figure \ref{fig:crazyflie_vel_icra} highlight the mission phase transitions and the rapid damping of capture transients.
It  shows velocity profiles (\(v_x, v_y, v_z\), and total speed). 
Distinct phase transitions (takeoff \(\rightarrow\) approach \(\rightarrow\) dock \(\rightarrow\) land) are visible, and docking transients are rapidly damped, yielding bounded and symmetric speeds through landing.

Overall, across simulation and hardware, the framework demonstrates (i) accurate relative tracking, (ii) robust stabilization under tight payload constraints, and (iii) repeatable midair docking behavior under progress-aware supervision. 
\subsection{Discussion, Limitations, and Future Directions}

\subsubsection{Discussion}

The results indicate that progress-aware phase gating and synchronized setpoint generation improve docking repeatability by reducing premature contact and limiting transient amplification during coupled leader-follower motion. The lightweight passive docking interface further increases engagement robustness by absorbing small residual pose errors at contact while remaining compatible with micro-UAV payload constraints.

\subsubsection{Limitations}

First, the current real-world evaluation relies on infrastructure-supported sensing (motion capture), which simplifies global pose estimation and time alignment. Deploying in uninstrumented environments will require onboard relative sensing (e.g., UWB/vision) and corresponding state estimation. Second, although magnetic latching improves repeatability, docking remains sensitive to closing speed and yaw misalignment, and systematic tuning and failure-mode analysis are therefore essential. Third, results are reported for the current docking geometry and safety constraints. Different connector designs or approaches can change success rates.

\subsubsection{Future work}

Future research will extend the platform in three directions: (i) introduce onboard relative sensing and time synchronization to reduce reliance on MoCap, (ii) expand benchmarking to a wider range of docking distances, approach speeds, and disturbance conditions, and (iii) port the supervision and evaluation pipeline to larger PX4-class UAVs to support contactless cooperative manipulation tasks.

\section{CONCLUSION}

This paper presented a reliable dual-drone midair docking platform for cooperative aerial manipulation under tight micro-UAV payload and thrust constraints. Two quadrotors operate in a leader-follower formation and execute midair docking using a lightweight modular frame with passive magnetic latching, supervised by a progress-aware mission supervisor that regulates phase transitions (approach, alignment, capture, and settle) and enforces tolerance-based guards to improve repeatability.
A complete hardware-software stack (ROS~2 with Crazyflie/PX4 interfaces) was implemented, enabling synchronized estimation, setpoint generation, and systematic evaluation. Simulation and indoor flight experiments demonstrate stable leader-follower docking behavior with bounded transients and consistent baseline/yaw regulation. The benchmark-style metrics reported in this work (success rate, time-to-dock, baseline/yaw consistency, and failure-modes) provide a practical foundation for statistically grounded comparison of docking supervision and controller configurations.
Future work will extend the platform toward operation in less-instrumented environments by incorporating onboard relative sensing and time synchronization, and will expand benchmarking across a broader range of docking distances, approach speeds, and disturbance conditions. We also plan to port the same supervision and evaluation pipeline to larger PX4-class UAVs to support contactless cooperative aerial manipulation tasks and longer-duration missions.

\addtolength{\textheight}{-12cm}   








\bibliographystyle{IEEEtran}
\bibliography{biblio}

\end{document}